\documentclass[letterpaper, 10 pt, conference]{ieeeconf}  

\IEEEoverridecommandlockouts                              

\usepackage{graphicx} 
\newcommand{\cmark}{\ding{51}}%
\newcommand{\xmark}{\ding{55}}%
\usepackage[hyphens]{url}
\usepackage{multirow}
\usepackage{pifont}
\usepackage{cite}

\title{\LARGE \bf
EmojiVoice: Towards long-term  controllable expressivity in robot speech\
}

\author{Paige Tutt\"os\'i$^{1,2,3}$, Shivam Mehta$^{4}$, Zachary Syvenky$^{1}$, Bermet Burkanova$^{1}$, Gustav Eje Henter$^{4}$ \\and Angelica Lim$^{1}$
\thanks{$^{1}$P. Tutt\"os\'i, Z. Syvenky, B. Burkanova and A. Lim are with the School of Computing Science,
        Simon Fraser University, 8888 University Dr., Burnaby, Canada
        {\tt\small ptuttosi@sfu.ca}}%
\thanks{$^{2}$P. Tutt\"os\'i is with SUPMICROTECH, CNRS, Institut FEMTO-ST, Université de Franche-Comté}%
\thanks{$^{3}$P. Tutt\"os\'i is with Enchanted Tools}%
\thanks{$^{4}$S. Mehta and G.E. Henter are with the the Division of Speech Music and Hearing, KTH Royal Institute of Technology}%
}

\begin{document}

\maketitle
\thispagestyle{empty}
\pagestyle{empty}

\begin{abstract}
Humans vary their expressivity when speaking for extended periods to maintain engagement with their listener. Although social robots tend to be deployed with ``expressive'' joyful voices, they lack this long-term variation found in human speech. Foundation model text-to-speech systems are beginning to mimic the expressivity in human speech, but they are difficult to deploy offline on robots.
We present EmojiVoice, a free, customizable text-to-speech (TTS) toolkit that allows social roboticists to build temporally variable, expressive speech on social robots. We introduce emoji-prompting to allow fine-grained control of expressivity on a phase level and use the lightweight Matcha-TTS backbone to generate speech in real-time. We explore three case studies: (1) a scripted conversation with a robot assistant, (2) a storytelling robot, and (3) an autonomous speech-to-speech interactive agent. We found that using varied emoji prompting improved the perception and expressivity of speech over a long period in a storytelling task, but expressive voice was not preferred in the assistant use case.

\end{abstract}

\section{INTRODUCTION}

Imagine a robot telling a 10-minute story to children. How would you like the robot to speak? The expression of paralinguistics such as emotions is an integral part of human speech \cite{beller10_speechprosody}, and humans convey expressivity by changing their expression over time \cite{clark2019makes, cowie2003describing}. Studies of expressivity in robot voices \cite{torre, robotlearn1, robotlearn2, robotlearn3, robotlearn4} use out-of-box voices such as those built into Nao and Pepper or commercial text-to-speech (TTS) such as Amazon Polly\footnote{https://aws.amazon.com/polly/}. These voices may contain, for example, one expressive ``joyful'' voice as an expressive voice. As a result, the use of a high-pitched and ``joyful'' voice\footnote{http://doc.aldebaran.com/2-4/naoqi/audio/altexttospeech-tuto.html?highlight=joyful} is common when attempting to give the impression of an expressive robot \cite{hennig2012expressive}, 
mirroring engaging speakers who use a higher-than-average pitch and greater pitch range \cite{NIEBUHR2016366}.


Nevertheless,  major challenges remain in current TTS systems for robots to mirror human vocal expressivity. First, although ``expressive'' and ``emotional'' TTS exist, they are difficult to deploy on robots; they do not work in real-time, have limited controllability, and are often not customizable. HRI researchers are not easily able to add their own expressive styles relevant to their use cases, nor do they have fine-grained control over a voice to explore specific hypotheses. Secondly, the use of ``expressive'' TTS in long-term interactions, such as reading a story, is poorly understood. As such, there is a need for a synthesized voice for robots that is not only ``joyful'' but temporally expressive, selecting the correct expressive style given the task and the linguistic context. We must, additionally, understand how and when different expressive styles should be deployed.

\begin{figure}[t]
  \centering
  \includegraphics[width=0.85\linewidth]{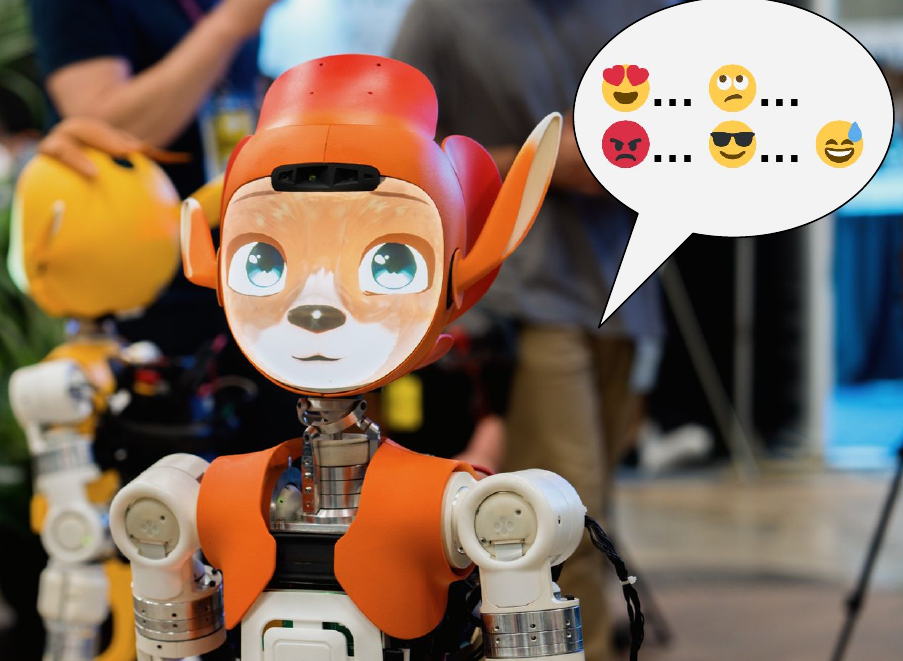}
  \caption{Miroka robot speaking with EmojiVoice expressive TTS.}
  \label{header}
\vspace*{-5mm}
\end{figure}

In this paper, we propose the use of emoji prompting to control long-term varied expressivity in robot speech (Fig. \ref{header}). Our contributions are as follows.
\begin{itemize}
    \item Introduce EmojiVoice, a free, open-source toolkit that enables social roboticists to create controllable, temporally variable, expressive voices for their robots using Matcha-TTS\cite{mehta2024matcha}.
    \item Propose emojis as a new means to control text-to-speech.
    \item Explore whether phrase-by-phrase emoji prompting can increase the perception of expressiveness.
    \item Investigate in which tasks a temporally variable expressive voice (vs singularly expressive voice) is effectively deployed.
\end{itemize}

\section{RELATED WORK}

\subsection{Controlling expressivity}
Many expressive TTS systems are controlled through a series of complex models and hours of expressive training data \cite{cho24_interspeech, bott24_interspeech}, which is contrary to the need for minimalist systems in robotics and is not flexible to fast fine-tuning to a specific use case. So-called ``expressive'' TTS systems mostly focus on increasing pitch and pitch range \cite{cho24_interspeech, bott24_interspeech}, as previous research has found that these modifications drive expressivity in human speech \cite{anotherstory, gustafson01_eurospeech}. Recent advancements in long-term TTS have improved smoothing over sentences by including contextual information from previous phrases \cite{xiao23_interspeech}. However, there is a lack of work exploring how this type of expressive voice performs over long periods of speech; for example, does listening to a singularly ``joyful'' expressive TTS for a long time become monotonous? 

One of the most related studies is a recent dataset released to help train Chinese expressive storytelling TTS through multiple annotations. They include emotional colouring, with the goal of creating a ``storytelling voice'' \cite{storytts}. They, however, do not provide an open-source model nor information on the model size, stability, and synthesis time of their baseline. Further, their use of an encoder to control emotion expressivity does not allow the user explicit expressive control, which has become a trend in TTS (especially in state-of-the-art paid solutions), where the expression is automatically selected based on text-based semantics and context\footnote{\url{https://www.hume.ai/blog/octave-the-first-text-to-speech-model-that-understands-what-its-saying}, \url{https://platform.openai.com/docs/models/gpt-4o-mini-tts}, \url{https://platform.openai.com/docs/models/gpt-4o-mini-tts}}, and may not be appropriate for HRI studies.

\subsection{Constraints of speech models in robotics}
Hardware capability continues to be a bottleneck in the explosive field of artificial intelligence (AI), where new, faster, and more efficient systems are often achieved at the expense of reliability, security, and cost \cite{gnad2024}. Many social robots deployed in HRI studies (e.g. Nao and Pepper: Aldebaran, United Robotics Group, iCub: Istituto Italiano di Tecnologia, LOVOT: GROOVE X, Furhat: Furhat Robotics) do not have an on-board GPU or TPU. A new generation of robots with GPUs are becoming available (e.g. Reachy: Pollen robotics, Mirokaï: Enchanted Tools). Yet, less than twenty of the 263 robots available on IEEE Robots database\footnote{\url{https://robotsguide.com/}} specify that they are equipped with GPU hardware. As such, AI models deployed on-board robots ideally have a small footprint and require a limited amount of resources at inference time \cite{aisize, distai}.

These hardware limitations are often circumvented through the use of cloud computing. Although this may be an adequate solution for in lab research deployment, often in-the-wild one cannot rely on a steady nor fast internet connection. This can lead to latency or even a complete inability to deploy interactive systems in-the-wild. In addition, the use of cloud resources can result in several security and privacy concerns \cite{privacy, MANIAH20191325}.

Lastly, social roboticists do not always come from an engineering background and may not have the domain knowledge required to deploy free-to-use TTS systems that often have poor documentation on their local deployment and customization. Because of this, HRI studies often use out-of-box paid solutions or those running on Hugging Face Spaces, which, due to shared GPU limitations, tend to not run in real-time resulting in the need to pre-render audio files, limiting the adaptability of the system. 

\begin{figure}[t]
  \centering
  \includegraphics[width=\linewidth]{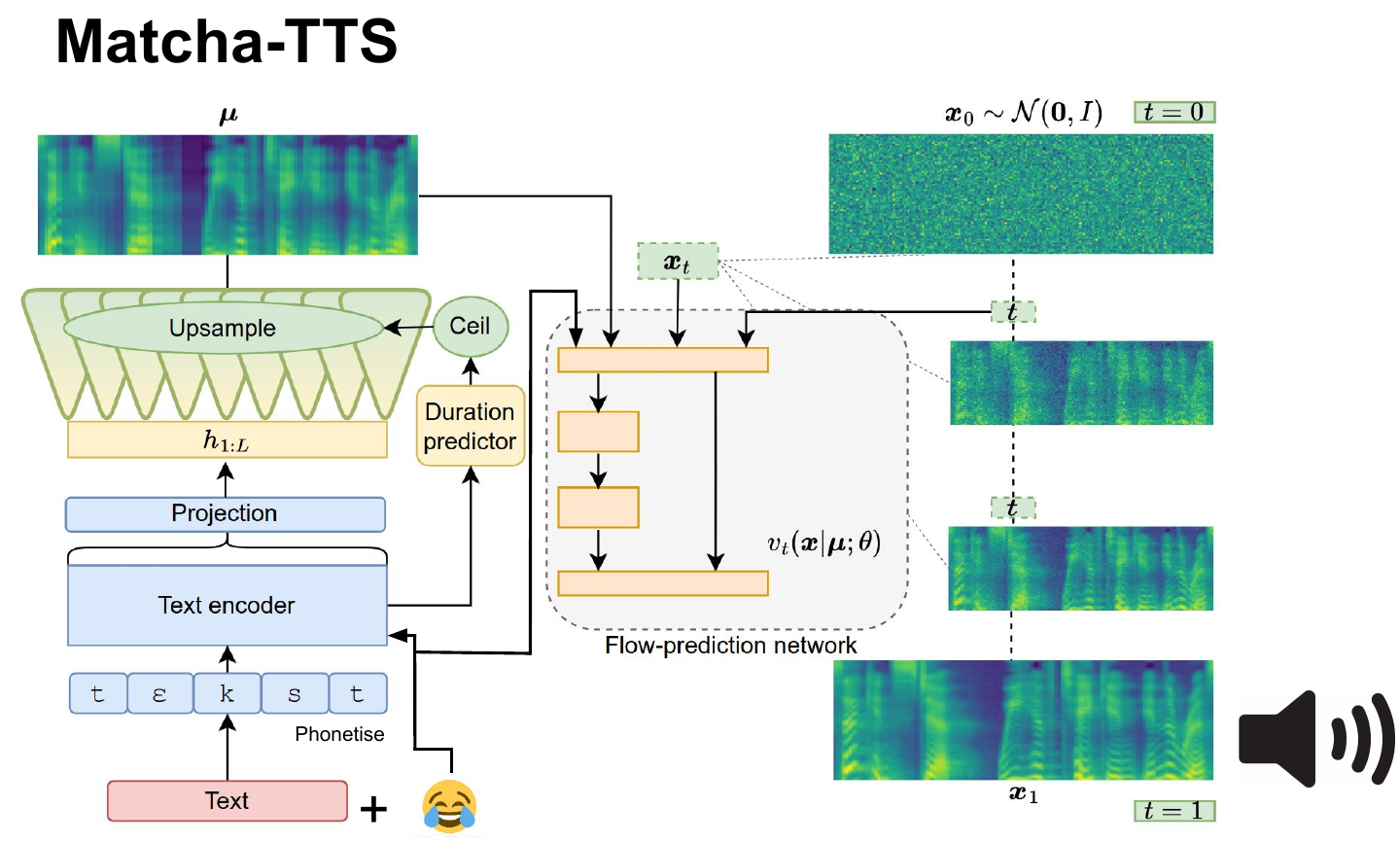}
  \caption{Matcha-TTS architecture \cite{mehta2024matcha} with the addition of emoji prompting to the text encoder and flow prediction network.}
  \label{arch}
\vspace*{-5mm}
\end{figure}

\begin{table*}[t]
\centering 
\caption{Comparison of open source TTS systems and robot-centric features}
\begin{tabular}{|p{3.2cm}|p{1.5cm}|p{1.5cm}|p{1.5cm}|p{1.5cm}|p{1.5cm}|p{2cm}|}
\hline
  & VoiceCraft \cite{peng2024voicecraft} & CoquiXTTS \cite{coqui} & Parler-TTS \cite{lyth2024naturallanguageguidancehighfidelity} & FastPitch \cite{fastpitch} & Matcha-TTS & Emoji-fine-tuned Matcha-TTS \\
 \hline
$<$ 100k parameters & \xmark & \xmark & \xmark & \cmark & \cmark & \cmark \\
No Hallucinations & \xmark & \xmark & \xmark & \cmark & \cmark & \cmark\\
real-time Synthesis  & \xmark & \xmark & \cmark & \cmark & \cmark & \cmark\\
Quick Fine Tuning*  & \xmark & \xmark & \xmark & \xmark & \cmark  & \cmark\\
Consistent Speaker Control & \xmark & \xmark & \xmark & \cmark & \cmark & \cmark\\
Controllably expressive  & \xmark & \xmark & \cmark & \cmark** & \xmark & \cmark\\
\hline
\end{tabular}
\label{compare}
\parbox{6in}{
\vspace{1mm}
\footnotesize{*on under 1 hour of fine-tuning data, **control pitch and duration by hand rather than intention or category}
}
\vspace{-5mm}
\end{table*}

\section{EMOJIVOICE TOOLKIT}\label{tts}
We propose a free, open-source toolbox with documentation that allows social roboticists to use emoji prompting to easily deploy an expressive TTS offline and customize the voices to their own use cases. 

\subsection{Text-to-speech (TTS) model}
We have adapted our text-to-speech model, Matcha-TTS, for the social robotics community. For the first time, we present a multi-speaker/voice version of Matcha-TTS \cite{mehta2024matcha}\footnote{\url{https://github.com/shivammehta25/Matcha-TTS/tree/main}} that allows for fast switching between emoji styles. Behind the scenes, Matcha-TTS is a neural TTS trained using optimal-transport conditional flow matching (OT-CFM). We have chosen this model for our toolbox as it is probabilistic rather than auto-regressive. Using a probabilistic model allows us to avoid hallucinations, i.e. ensuring the output audio will always match the input text, as well as issues with speaker identity control. Moreover, the model is ideal for robot's applications as it has a compact memory footprint of only 20.9M parameters, in comparison to 830M for VoiceCraft and 41.2M for Fastspeech 2, and has a real-time-factor (RTF) of 0.3 (an RTF $<1$, taking at most 1 second to generate 1 second of audio, is considered real-time \cite{pratap20b_interspeech}) on a Nvidia Jetson AGX Orin 64GB, a GPU with similar power and memory to those applied in production robotics. Multi-speaker Matcha-TTS takes a label, in the case of the EmojiVoice toolbox an emoji number, along with the input text. This is concatenated both to the text encoder input and the flow-prediction network (decoder) input in order to contain multiple voices in a single, small (78MB, in our case) checkpoint. The architecture can be seen in Fig. \ref{arch}. Having multiple voices in a single checkpoint both reduces space and allows for efficient switching between voice styles without needing to reload the model.

\subsection{Emoji Representation}
\begin{figure}[t]
  \centering
  \includegraphics[width=\linewidth]{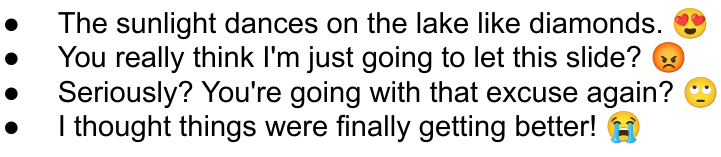}
  \caption{Example prompts for voice actors for data collection and training.}
  \label{prompts}
\vspace{-7mm}
\end{figure}

\begin{figure}[t]
  \centering
  \includegraphics[width=\linewidth]{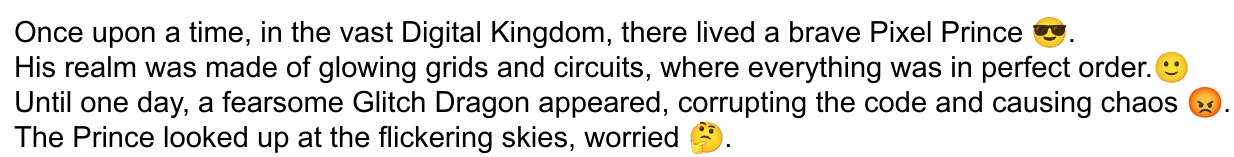}
  \caption{Example of text with emojis. Sample of the script provided to the TTS for Case Study 2: Storytelling task.}
  \label{script}
\vspace*{-3mm}
\end{figure}

We suggest emojis as a means to prompt expressive styles that can be used for temporal variational control, i.e., selecting a different emoji style for each phrase generated. In 2019, \cite{bai2019} stated that there were 3,019 emojis in Unicode, and this number has now grown to 3,782 in only 5 years. While not all emojis can be considered representations of human expression, considering 106 emojis are smileys alone, with the addition of non-smiley emojis such as ``thumbs up'' and ``flexed biceps,'' we have a rich representation of both emotions and social signals larger than, or at least as large as the most commonly used discrete emotion models \cite{schroder2006first, ShaverPhillip1987EKFE, PLUTCHIK19803}.  The variation provided by the standard Unicode emoji set is especially useful in that they can express emotions: \includegraphics[scale=0.3]{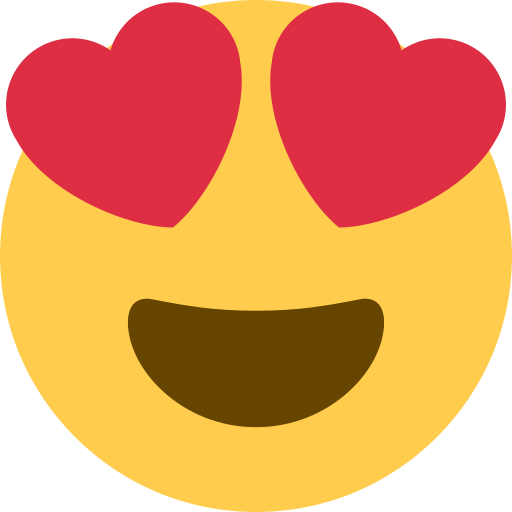}, \includegraphics[scale=0.018]{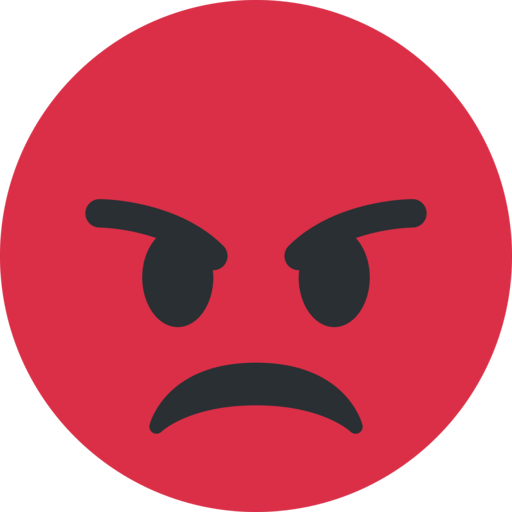}, attitudes: \includegraphics[scale=0.04]{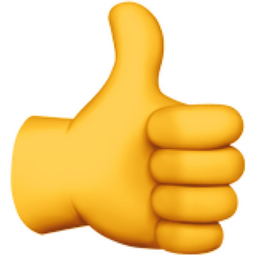}, \includegraphics[scale=0.018]{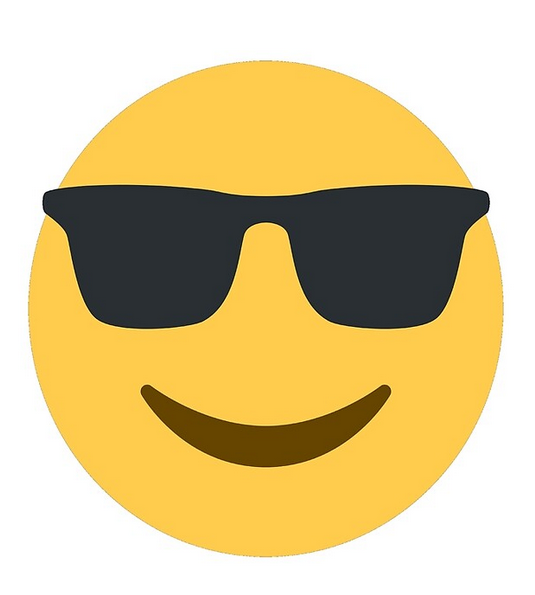}, mental states: \includegraphics[scale=0.018]{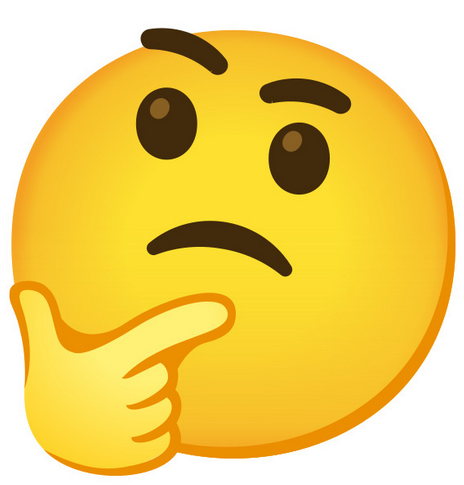}, \includegraphics[scale=0.018]{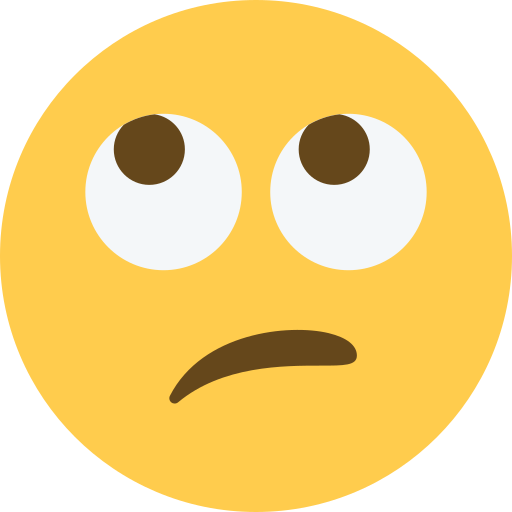}, and bodily states: \includegraphics[scale=0.1]{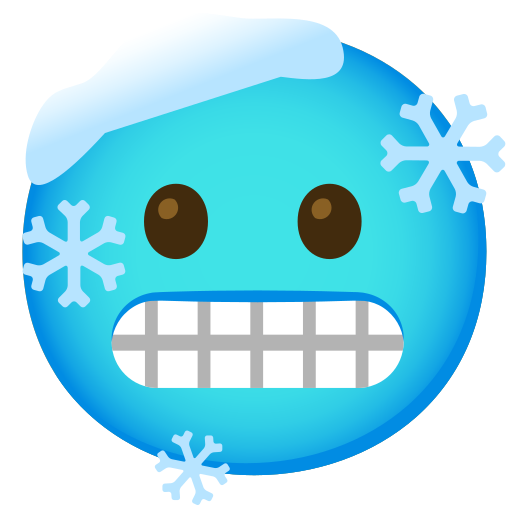}, \includegraphics[scale=0.005]{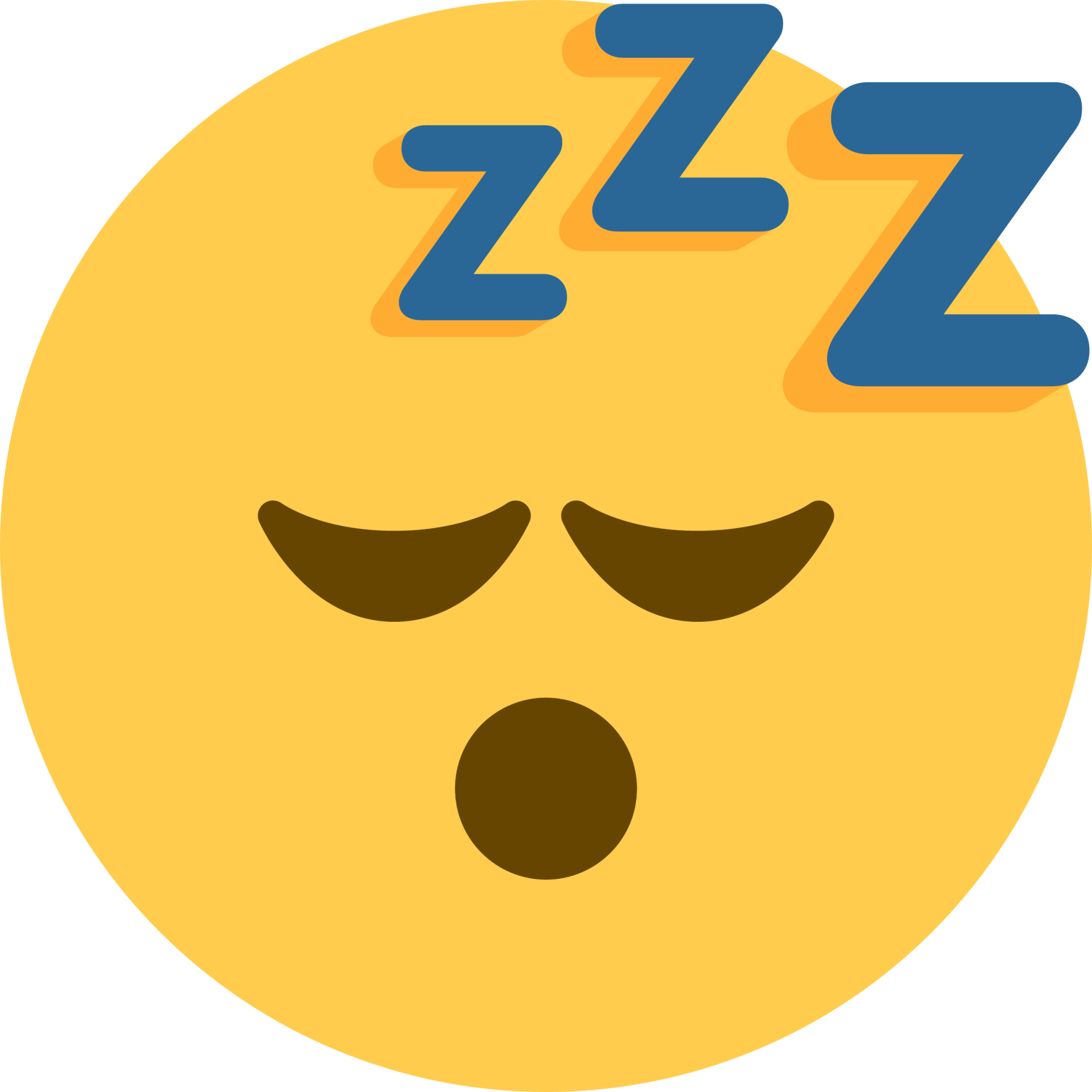}. Moreover, we have become accustomed to the use of emojis in our everyday written expression; for example, almost half of all Instagram posts with text contained an emoji in 2015 \cite{dimson2015emojineering}. This has resulted in a large amount of the training data for large language models (LLMs) containing emojis along with the text; hence, LLMs can readily produce emojis within text in a rather convincing manner \cite{savage2024real, dunngood}. Although LLMs are large and costly, they are increasingly incorporated into robotic conversational systems and are a promising option for generating dialogue~\cite{10611232, 10599883, ZHANG2023100131, 10.1145/3568294.3580040}. This means that emoji-based voice selection can be easily integrated into systems using an LLM with a negligible added cost. Thus, we used emojis to 1) collect emotional vocal expressions in humans to generate training data for our expressive TTS (Fig. \ref{prompts}) and 2) control the TTS voice selection on a phrase-by-phrase basis to ensure long-term temporal variability as seen in Fig. \ref{script}.

\begin{figure}[t]
  \centering
  \includegraphics[width=\linewidth]{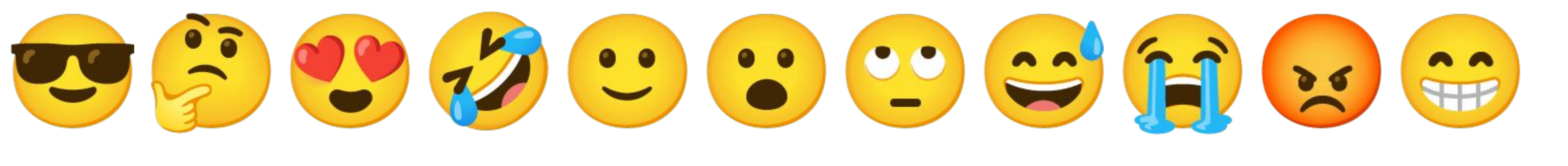}
  \caption{Emojis used for voice prompting and voice selection.}
  \label{emojis}
\vspace*{-7mm}
\end{figure}

The voices available in EmojiVoice were fine-tuned on the data of three speakers (Paige, Olivia, Zach) prompted with 11 different emojis (Fig. \ref{emojis}). To select the emojis, we aimed to have more positive than negative emojis as this is more appropriate for social robots. We do not claim that the present set of emojis is the optimal set; rather, it is a starting point for researchers to explore, expand or even limit the expressivity of robots using this representation, and researchers can use as little as 3 minutes of speech per emoji to customize the TTS. A comparison of Matcha-TTS and the emoji-fine-tuned Matcha-TTS with other freely available, well documented, off-the-shelf TTS systems can be seen in Table \ref{compare}.

\subsection{Data Collection and Training}
We provide information on data collection for other researchers to extend or retrain EmojiVoice depending on their needs. To collect training data, we created a prompting script for the speaker. We generated 50 sentences for each emoji using ChatGPT4-o\footnote{https://openai.com/} with the prompt: \emph{Please provide 50 short phrases that reflect this emoji: X. The phrases should use all the English phonemes and should not be repetitive, using a variety of words.} These phrases appeared on the screen one at a time with the target emoji appended to the end of the sentence, and the speaker pressed a button each time they were ready to record. Example phrases can be seen in Fig. \ref{prompts}. The data was split into 40 sentences of training data and 10 for validation. We fine-tuned off of the multi-speaker VCTK checkpoint available for Matcha-TTS\footnote{\url{https://drive.google.com/drive/folders/17C_gYgEHOxI5ZypcfE_k1piKCtyR0isJ}}. We trained on an NVIDIA GeForce RTX 4070 GPU with an Adam optimizer and a learning rate of 1e-4, unchanged from the checkpoint, with a batch size of 20. The model was trained for 85 epochs, approximately 20 minutes. We trained three checkpoints from three different actors, two female and one male, using our toolbox. The toolbox extends Matcha-TTS for ease-of-use by HRI researchers by including: 1) Training files setup: examples, raw data, and 3 checkpoints (with and without optimizers), 2) Additional information on the amount of data needed to fine-tune, 3) Scripts to record the data, 4) Wrappers to parse emojis in text to prompt the voices, and 5) A conversational agent. Our toolbox, data, and voices are available free and open source\footnote{\url{https://github.com/rosielab/emojivoice}}.

\subsection{Conversational Agent}\label{s2s-model}
We include an autonomous, speech-to-speech interactive agent chaining automatic speech recognition (ASR), an LLM and our TTS with emoji selection to create an autonomous dialogue system, which we explore in Case Study 3 (section \ref{s2s}). There may remain some latency issues in the LLM on low-power GPUs, but for our testing in Case Study 3 (RTX 4060) it ran in real-time. We use OpenAI's Whisper \emph{tiny.en} model \cite{whisper2023} for speech recognition (39M parameters). To mitigate the complexity of deciding when the user has finished speaking \cite{speakingturn}, we use a push-to-talk system, recording the voice and producing a transcription. The transcription is the input to the LLM, currently Meta's Llama3.2 \cite{dubey2024llama3herdmodels} (1.23B parameters) implemented via Ollama\footnote{https://ollama.com/} and LangChain\footnote{https://www.langchain.com/} as a chatbot. In the toolbox, the voice designer is able to control the 1) prompt to the LLM to a specific use case and 2) emoji set. This prompt controls the concatenation of the emoji to each response. We extract the appended the emoji and use a mapping from the emoji to the associated voice numbers, providing this to the TTS to synthesize the appropriate voice.

\begin{figure}[]
    \centering
    \begin{minipage}{0.49\linewidth}
  \includegraphics[width=\linewidth]{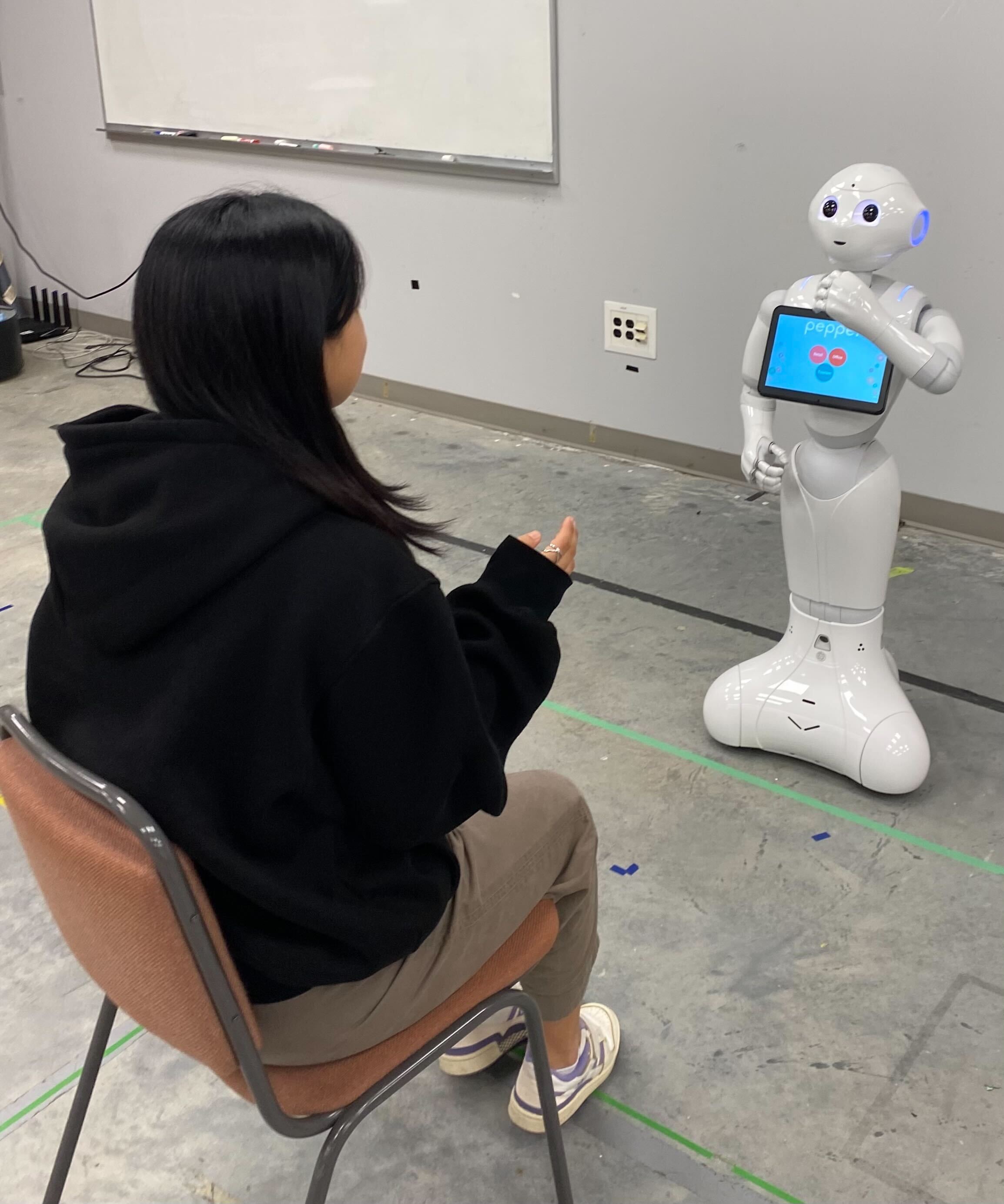}
  \caption{Case Study 1 and 2: participant view of the interaction with a Pepper robot.}
  \label{setup2}
  \vspace*{-4mm}
    \end{minipage}%
    \hfill
    \begin{minipage}{0.49\linewidth}
  \includegraphics[width=\linewidth]{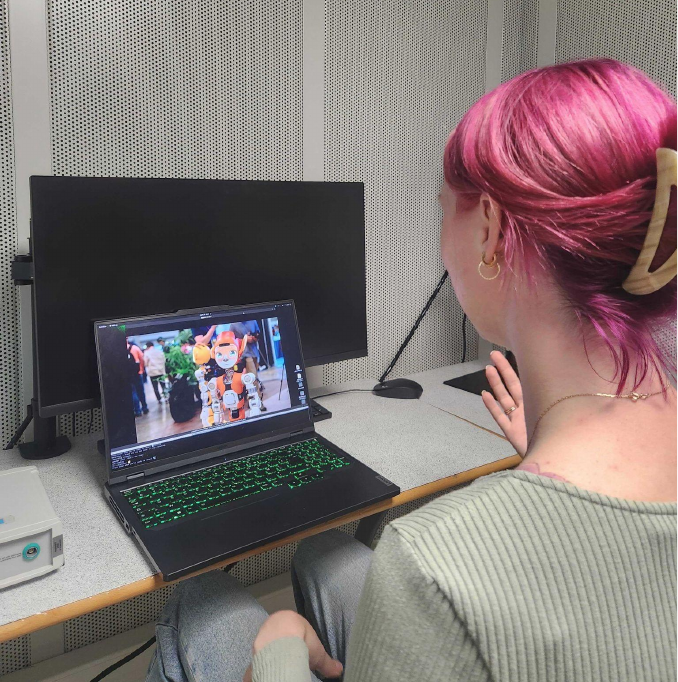}
  \caption{Case Study 3: Participant interacting with the storytelling autonomous speech-to-speech and an image of a Miroka robot.}
  \label{setup3}
    \end{minipage}
\vspace{-3mm}
\end{figure}

\section{CASE STUDIES}
We explore three possible use cases for the voices available with EmojiVoice to understand how an expressive voice can be effectively deployed in HRI and if the use of per-phrase emoji prompting can increase the perception of expressiveness in a voice over a ``joyful'' model. The first case is a scripted conversation with a robot; the second is a short story told by the robot; and the third is an autonomous interactive agent playing a storytelling game with the user. The first two case studies were conducted both on the Pepper and the Miroka (Enchanted Tools) robots, and the third was with an image of a Miroka robot. Each case study compared three voices (using the ``Paige'' voice): 1. \textbf{Baseline}: the original voice from the Matcha-TTS VCTK checkpoint for speaker one for all phrases; 2. \textbf{Pleasant}: the EmojiVoice voice prompted with only the \emph{slightly smiling} emoji, which has an increased pitch range over the baseline representing a ``joyful'' voice that is often employed in expressive TTS systems for robotics; 3. \textbf{Emoji}: the full 11 set of emoji voices with each phrase containing an emoji to select a specific voice. Both the baseline voice and training data for the EmojiVoice voices were of speakers perceived to be 25-35 years old and female. The study received internal ethics approval from the REB at Simon Fraser University.

\subsection{Case Study 1: Conversational Robot Helper}

\begin{figure}[t]
  \centering
  \includegraphics[width=\linewidth]{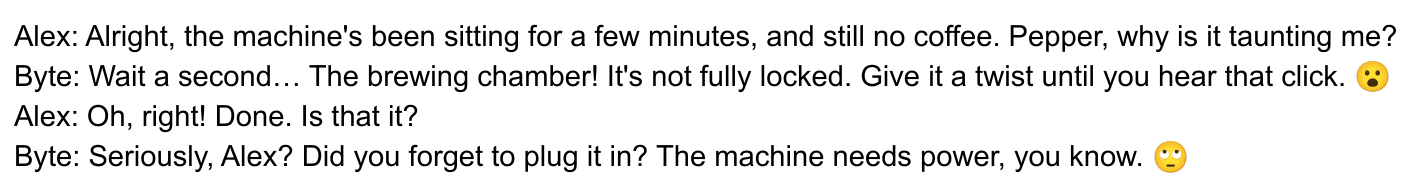}
  \caption{Case Study 1: Robot Helper with human ``Alex'', script excerpt.}
  \label{case1script}
  \vspace*{-3mm}
\end{figure}

\begin{figure}[t]
  \centering
  \includegraphics[width=\linewidth]{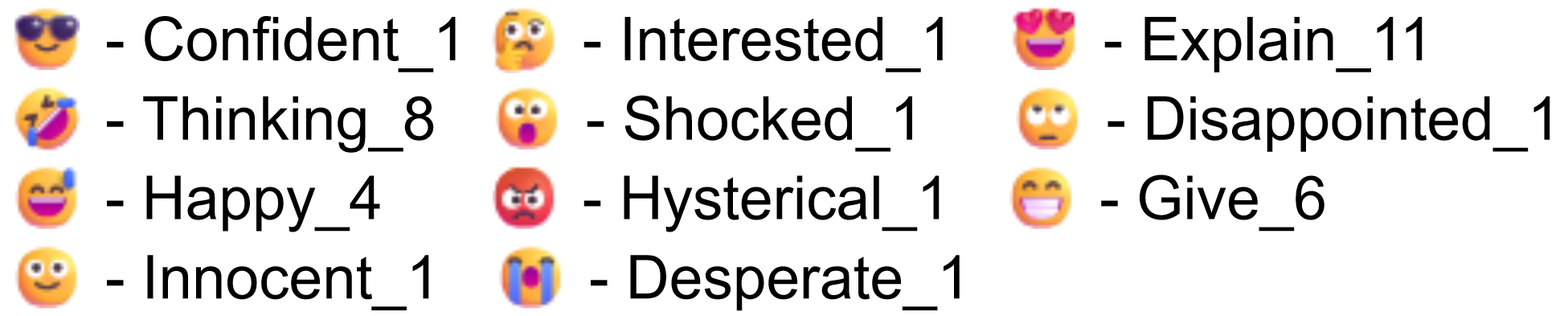}
  \caption{Animations played on Pepper in Case Study 1 for each emoji}
  \label{emojis and accompanying animatoins}
\vspace*{-7mm}
\end{figure}

\subsubsection{Preparation and setup}\label{setup}
For our first case study, participants observed a short, scripted conversation. This case allows us to investigate the perception of short sentences and the interplay with human dialogue.

The script for the conversation was generated with ChatGPT4-o using the prompt: \emph{Can you please generate interaction between two people, a human and a robot helper, where the robot's lines are always coloured with one of these emojis? XXXXXXXXXXX Each of the emojis must be used at least once. It should be about a 5 minute conversation.} The snippet of the script can be seen in Fig. \ref{case1script}.

For the Pepper study the voice was played along with a set of animations\footnote{http://doc.aldebaran.com/2-5/naoqi/motion/alanimationplayer-advanced.html\#animationplayer-list-behaviors-pepper} (Fig.~\ref{emojis and accompanying animatoins}). These were chosen to both match the expressivity of the emoji for each phrase and to be short enough to not cause a lag in the conversation. Basic awareness was disabled to avoid the robot randomly looking around when speaking, which can affect consistency across conditions. For the Miroka experiment, facial animations matching the selected emojis accompanied the voice, and lip sync was on. For both robots, the animations were the same for all of the voices. The participants were aware that this conversation was not autonomous, and both the robot and the sound were being controlled by the researchers. The order of the voices were: Baseline, Emoji, Pleasant.

\begin{figure}[t]
  \centering
  \includegraphics[width=1.03\linewidth]{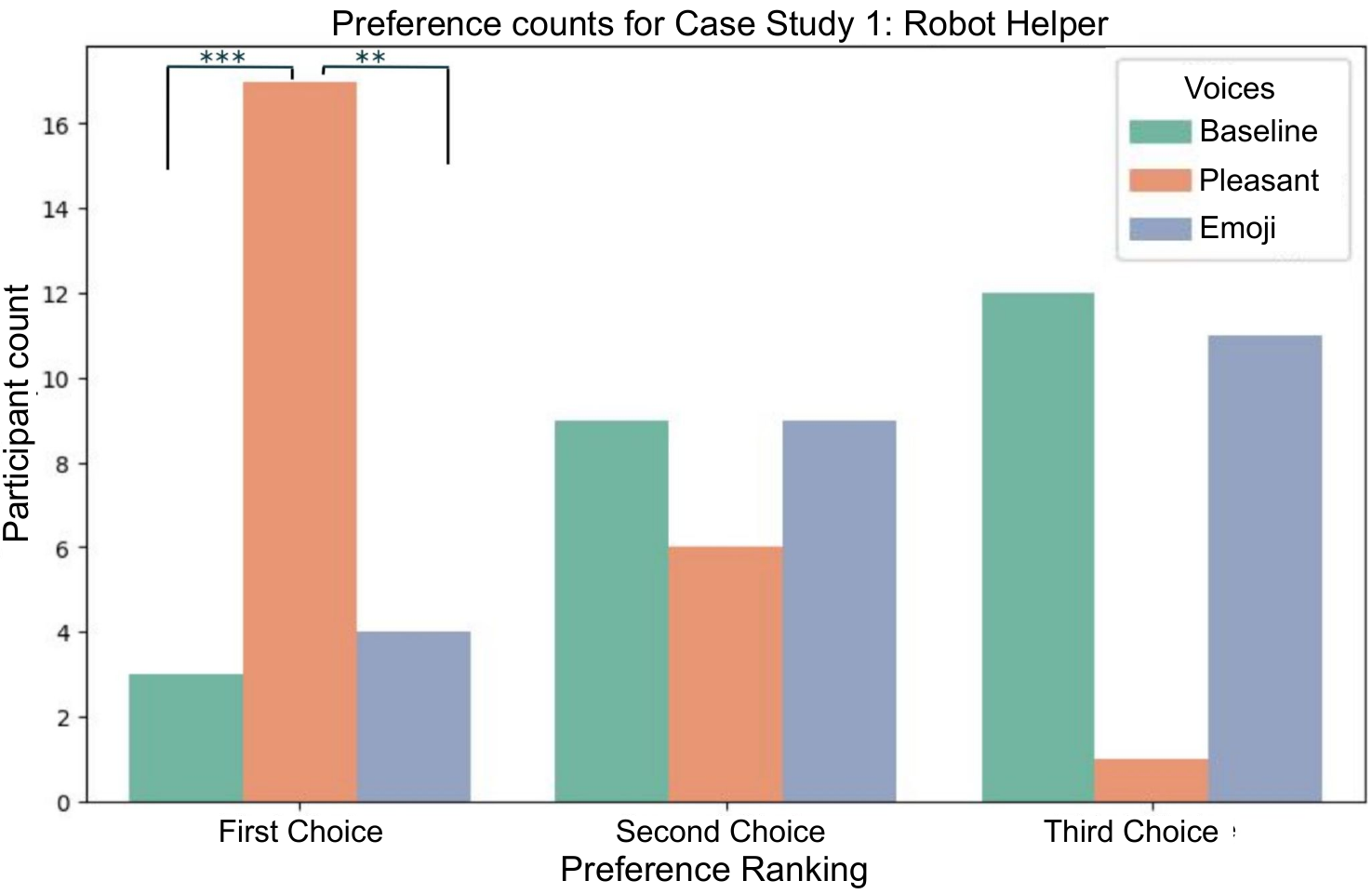}
  \vspace*{-5mm}
  \caption{Case Study 1: Robot Helper. Counts for voice preference}
  \label{case1pref}
\vspace*{-7mm}
\end{figure}

\begin{table*}[t]
 \hspace*{-2mm}
 \caption{Case Study 1: Results for Difference of Mean Statistical Tests}
 \vspace{-2mm}
 \label{case1anova}
\begin{tabular}{ |p{1.1cm}|p{0.4cm}|p{0.4cm}|p{0.4cm}|p{0.4cm}|p{0.4cm}|p{0.4cm}| p{0.4cm}|p{0.4cm}|p{0.6cm}|p{1.9cm}|p{4.8cm}|p{0.5cm}| }
 \hline
\multirow{2}{*}{} &
  \multicolumn{2}{c|}{Baseline} &
  \multicolumn{2}{|c|}{Pleasant} &
  \multicolumn{2}{|c|}{Emoji} &
  \multirow{2}{*}{DF} &
  \multirow{2}{*}{N} &
  \multirow{2}{*}{F} &
  \multicolumn{2}{c|}{P-value} &
  \multirow{2}{*}{$\omega^2$}\\
  
   & $\mu$ & $\sigma$ & $\mu$ & $\sigma$ & $\mu$ & $\sigma$ & & & & One-Way Anova & Tukey HSD &\\
 \hline
 xMOS   & 1.71 & 1.0 & \emph{5.71} & \emph{2.03} & \textbf{7.42} & \textbf{1.91} & 2 & 24 & 70.46 & $<0.001$*** & $<0.001$*** (E-P, P-B), 0.003** (E-P) & 0.67\\
 sMOS  & 3.58 & 2.26 & \emph{6.63} & \emph{1.81} & \emph{5.96} & \emph{2.22} & 2 & 24 & 13.81 & $<0.001$*** & $<0.001$***(P-B), 0.001**(E-B) & 0.26\\
 Suitability  & 4.54 & 2.84 & \textbf{6.96} & \textbf{2.33} & 5.25 & 2.09 & 2 & 24 & 6.21 & 0.003** & 0.003**(P-B), 0.04*(P-E) & 0.13\\
 \hline
 
\end{tabular}
{\parbox{6.8in}{
\footnotesize 
1. \textbf{Bold} values highlight the voice that is significantly higher than both other voices, \emph{italics} indicate it is significantly higher than one other voice\\
2. In the Tukey HSD the results are presented as (higher rated voice - lower rated voice) by the first letter of the the voice name\\
3. Due to space limitations, we only report those voices that were found to have significant differences.
}
}
\vspace{-5mm}
\end{table*}

Participants were seated in a group behind the user performing the interaction, as can be seen in Fig. \ref{setup2}. Following the interaction with each of the voices, the participants filled out a survey. We used MOS scores for prosody (pMOS), intelligibility (iMOS), and social impression (sMOS) from the MOS-X2 scale \cite{Lewis2018InvestigatingMR}, and for expressiveness of intonation (xMOS) as in \cite{korotkova24}. Lastly, we questioned the suitability of the voice for the robot following \cite{rijn2024}. We used a 10-point Likert scale for all ratings following the MOS-X2 scale to maintain consistency across the questions. Once all three voice interactions had been completed, the participants ranked their preferred voice and explained their choice. Finally, there was an open group discussion to brainstorm and discuss the perception of the voices.

\subsubsection{Participants}\label{participants}
There were 24 participants total (10F, 14M; 16 Pepper, 8 Miroka). There were 16 participants aged 25-35, 4 aged 35-45, and 4 aged 18-25. The most prominent ethnicity was White (10), followed by Chinese (8), then West (4), Southeast (1) and South Asian (1). Six of the participants had English as a first language, but 11 of them use English as their most common day to day language (French, Italian, Chinese and Persian being the others). The participants reported a self rated English proficiency on a scale of 1-5, two participants rated their proficiency as a 3, twelve as a 4 and ten as a 5. All participants completed a verbal consent process as part of ethical requirements, and no personally identifying information was collected.

\subsubsection{Results}\label{results1}
We used a one-way ANOVA (Type II) followed by a post-hoc Tukey HSD and set $\alpha=0.05$ to assess differences in the opinion ratings between voices. These results can be found in Table \ref{case1anova}.

We found that the Pleasant voice was significantly more suitable for the robots (no difference between robots) than both the other voices. Both the Emoji and Pleasant voice were found to have a significantly higher social impression than the Baseline voice. Lastly, the emoji voice was found to have a significantly higher expressivity than both other voices, and Pleasant was more expressive than the Baseline.

We completed Chi Squared Goodness of Fit tests, to assess for voice preference, followed by a bootstrapped confidence intervals (CI)(Fig. \ref{case1pref}). We found that the Pleasant voice was preferred as the first choice over the Emoji voice at a 99\% CI, and over the Baseline voice with over a 99.9\% CI.

The discussion of this case study yielded interesting results. Although the Baseline voice is neutral read speech the participants agreed that the voice sounded ``tired and depressed'' or ``apathetic'' and one participant suggested that perhaps a neutral voice would have been better for the study. This same commentary was lacking in further case studies using this voice, where it was indeed perceived as neutral. In addition, the participants noted for the Emoji voice that, ``it sounded a bit sassy ... a bit too much in the emotions, like a robot that is judging you.'' These comments may be a result of the text and forcing the LLM to provide emojis to a robot assistant conversation, resulting in an overly dramatic script. In this case, the participants noted that the pleasantness of the classically ``expressive'' Pleasant voice made the robot sound ``just the right amount of frustrated, like it was annoyed with the task but it still wanted to keep helping you.''

\subsection{Case Study 2: Storytelling}

\begin{table*}[t]
 \hspace*{-2mm}
 \caption{Case Study 2: Results for Difference of Mean Statistical Tests}
 \vspace{-2mm}
 \label{case2anova}
\begin{tabular}{ |p{1cm}|p{0.75cm}|p{0.75cm}|p{0.75cm}|p{0.4cm}|p{0.4cm}|p{0.55cm}| p{0.55cm}|p{0.55cm}|p{0.88cm}|p{2.0cm}|p{3.0cm}|p{0.75cm}| }
 \hline
\multirow{2}{*}{} &
  \multicolumn{2}{c|}{Baseline} &
  \multicolumn{2}{|c|}{Pleasant} &
  \multicolumn{2}{|c|}{Emoji} &
  \multirow{2}{*}{DF} &
  \multirow{2}{*}{N} &
  \multirow{2}{*}{F} &
  \multicolumn{2}{c|}{P-value} &
  \multirow{2}{*}{$\omega^2$}\\
  
   & $\mu$ & $\sigma$ & $\mu$ & $\sigma$ & $\mu$ & $\sigma$ & & & & One-Way Anova & Tukey HSD &\\
 \hline
 xMOS   & 2.95 & 1.57 & 4.04 & 2.01 & \textbf{7.75} & \textbf{1.98} & 2 & 24 & 43.49 & $<0.001$*** & $<0.001$***(E-P, E-B) & 0.54\\
 sMOS   & 3.67 & 2.01 & 4.96 & 1.78 & \emph{6.29} & \emph{2.26} & 2 & 24 & 10.07 & 0.001** & $<0.001$***(E-B) & 0.20\\
 Suitability   & 4.17 & 2.51 & 5.29 & 2.42 & \emph{6.08} & \emph{2.41} & 2 & 24 & 3.71 & 0.030* & 0.023*(E-B) & 0.13\\
 \hline
\end{tabular}
\vspace*{-5mm}
\end{table*}

\subsubsection{Preparation and set up}
Next, we considered reading a short story. This case allows us to investigate the perception of the voice when stringing together several different sentences without interruption. In particular, we suspect that variation in the expressions is an important factor, as speaking continuously with the same ``expressive'' voice can either become boring or annoying over time. Moreover, the story was generated one sentence at a time in order to test the real-time nature of the TTS.

The story was generated with ChatGPT-4o using the prompt: \emph{can you please tell me a short story using as many of these emojis as possible XXXXXXXXXXX. The emojis should reflect the emotion in the voice when reading the text and not a physical action. The story should be a fairy tale in a digital realm. I want it to take about 1 or 2 mins to read out-loud.} A sample of the script can be seen in Fig. \ref{script}.

The same setup was employed for the robot. This time, grand gesture animations appropriate for storytelling were chosen for Pepper (Everything\_4, Far\_3, Thinking\_8, ShowSky\_11). Facial animations were used similarly for Miroka. The order of the voices were: Pleasant, Baseline, Emoji.

The setup was the same, but this time the story, without the accompanying emojis, was projected for the participants to read along. This was done to mimic a storybook reading scenario. Following the interaction with each of the voices, the users filled out the same survey and completed discussion as in Case Study 1, see Sec. \ref{setup} for further information.

\subsubsection{Participants}
Case Study 2 used the same participant pool as Case Study 1, see Sec. \ref{participants}.

\subsubsection{Results}\label{results2}
\begin{figure}[t]
  \centering
  \includegraphics[width=1.03\linewidth]{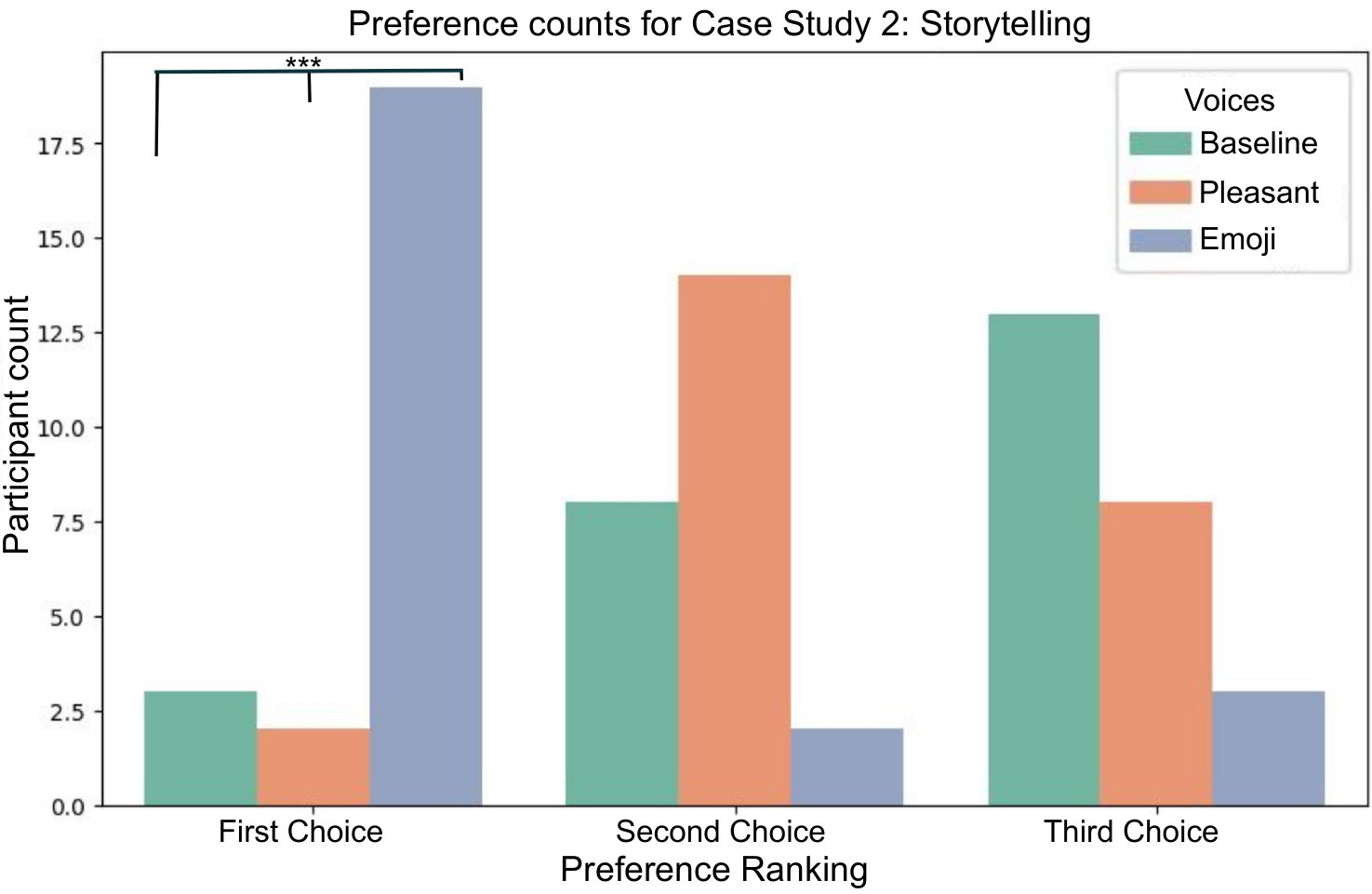}
  \vspace*{-2mm}
  \caption{Case Study 2: Storytelling. Counts for voice preference}
  \label{case2pref}
\vspace*{-5mm}
\end{figure}

The same assessment was used as in Case Study 1; see Sec. \ref{results1}. The results can be found in Table \ref{case2anova} and a visualization of the preference counts in Fig. \ref{case2pref}.

We found that the Emoji voice was significantly more expressive than both the Baseline and Pleasant voice. We also found the Emoji voice to have a significantly higher social impression and suitability than the Baseline voice (no significant difference between the robots). The emoji voice was preferred as the first choice voice over the other two voices at a 99.9\% CI.

This time, during open discussion, the opinion was strongly towards the need for a highly expressive voice for a storytelling task. Participants said ``you can never be too expressive in storytelling,'' ``the \{Emoji voice\} is the most expressive, which is very important in storytelling,'' ``I feel that the robot should express emotion according to the story content,'' and ``the \{Emoji voice\} would be the best for children because of the sudden changes in all aspects of the voice.'' Additionally, one participant mentioned, ``\{Pleasant voice\} was interesting at first but became boring over time,'' supporting our hypothesis that over time, a voice that is expressive line by line (in Case Study 2) becomes boring for long text. Yet, there were still some comments on inconsistency in the expression, ``\{Emoji voice\} was entertaining but sometimes the timing, pitch, and emphasis felt off.'' This may suggest that we need an even lower, more granular level of control over expressivity (see Sec. \ref{discuss}).

\subsection{Case 3: Autonomous speech-to-speech interactive agent}\label{s2s}

\subsubsection{Preparation and set up}
\begin{figure}[t]
  \centering
  \includegraphics[width=1.03\linewidth]{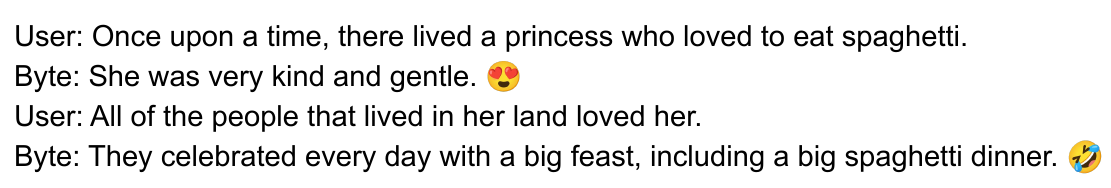}
  \vspace*{-5mm}
  \caption{Case Study 3: Story Building, example interaction.}
  \label{case3script}
  \vspace*{-5mm}
\end{figure}

For our final case study, we wanted to better understand how participants view the voices in an autonomous, interactive scenario with the voice. To this end, Case Study 3 was a one-on-one interaction directly between the participant and the system. The exact prompt for the LLM can be found along with the toolbox. The LLM was asked to play a game where it constructs a story by taking turns with the user. The LLM was additionally told to append an emoji to the end of every response it provided that reflects the expression of the phrase. The users played until 2 minutes had passed. As we had a new set of participants and they were interacting one at a time, the order of the voices was selected pseudo-randomly (random while still ensuring all orders were covered) for each participant. The participants sat in front of a computer screen, as seen in Fig. \ref{setup3}, to complete the interactions. For this more complex, highly divergent interaction where several factors can contribute to participant perceptions, we opted for only a qualitative discussion and assessment of the voices. An example script can be seen in Fig. \ref{case3script}

\subsubsection{Participants}
There were 8 participants in total (5M, 3F). There were 5 participants aged 25-35, 2 aged 18-25 and one aged 35-45. The most prominent ethnicity was Arab (3), followed by Caucasian (2) and South Asian (2) and one West Asian participant. Only 1 of the participants had English as a first language, but 3 of them use English as their most common day-to-day language (French, Arabic, and Chinese being the other three). The participants reported a self-rated English proficiency on a scale of 1-5; one participant rated their proficiency as a 3, two as a 4 and five as a 5. All participants completed a verbal consent process as part of ethical requirements.

\subsubsection{Results}
As each user interaction was different due to the unique improvisation of the user and system, we opted to only complete an open interview for this case study. From participant discussions, we learned that although the participants were aware that the overall goal of the study was to assess the voice, they found their motivations for engagement were highly motivated by the LLM and the setting of the game rather than the voice. One participant said, ``It was really fun, but the reason I wanted to keep going was I thought the game was funny, I didn't really consider the voice.'' Moreover, the performance of the LLM, specifically in its ability to choose emojis, heavily influenced participant perception of expressivity and social impression. One participant said, ``It was saying that the princess fell deeper into addiction, but it said it with a laughing voice... it was funny but really inappropriate,'' giving a negative perception due to a poorly selected emoji. Whereas another noted, ``there was an overly sad tone on one sentence (something about piglet for which she was clearly devastated), which created a sense of irony and playfulness,'' where the emoji was selected appropriately and therefore led to a more positive impression of the voice.  The participants did, however, want to hear the robot be expressive: ``I expected to hear more expressions when they were talking,'' speaking of the Baseline voice. Lastly, the participants noted that the system felt like it was in real-time, and any lag seemed appropriate for the robot to be thinking of the next sentence.

\section{DISCUSSION, LIMITATIONS, AND FUTURE DIRECTIONS}\label{discuss}

A thorough overview of the struggles of designing voices for robots can be found in \cite{MARGE2022101255}. There are 25 recommendations organized into 7 categories: user experience design; audio processing, speech recognition, and language understanding; speech synthesis and language generation; dialogue; other sensory processes; robustness and adaptability; and infrastructure. In our work, we addressed their suggestions for user experience, speech synthesis and, language generation and dialogue.

We aimed to meet their suggestions in user experience by reporting on three different possible use cases where we expect the needs for a robotic and expressive voice will differ and explored where and what type of expressive voice should be applied. We also focused on expressivity rather than naturalness to maintain a robotic voice in an attempt to manage expectations. We supported the recommendations for speech synthesis and language generation by using the lightweight and efficient Matcha-TTS fine-tuned to a small but specific subset of data and by using emojis to \emph{``Extend the pragmatic repertoire of speech synthesizers.''}. Lastly, we addressed recommendations for dialogue through the use case of engaging dialogue with a speech-to-speech system that runs in real-time.

\subsection{Discussion}
When comparing across the three cases, we see that the Emoji voice was most appropriate for an expressive, longer-term interaction such as storytelling. This is an interesting result since with TTS models, there is a tendency to avoid using long blocks of text; the longer you hear a TTS, the more monotonous it may sound \cite{neural-tts-long-form}. Specifically, the expressivity for the Pleasant voice decreased when used in a long-term rather than line-by-line interaction. Remarkably, by selecting a different emoji for each phrase at run-time, we were able to lessen this effect.

Additionally, we found that the perception of a voice can completely change depending on the context of the task and script. Although the same voices were used between the first two case studies, in the first study, participants reported that the Baseline voice sounded depressed and like it didn't want to help the user, whereas it sounded merely neutral and monotonous when telling a story. The Emoji voice seemed judgmental and sassy in the first study, compared to just being expressive in the second. In the case of an assistant, it appears that simply choosing a more pleasant voice allowed the participants to feel like the robot was more willing to help with the task, and highly expressive voices are not needed, compared to the storytelling where the participants expect a high degree of expressive variability over the phrases to keep their interest. This may also be due to forcing emojis such as ``eye rolling'' and ``anger'' into the assistant task script. It remains future work to disentangle the effect of general expressivity and selection of the appropriate expressivity for the social context.

Both through the data recording process and the case studies, we note that although the system seems to be appropriate for long-form interactions with variations from phrase to phrase, we still observe awkwardness in very long phrases. Indeed, humans often will use multiple expressive tones over the course of a long phrase. Our system is limited to one emoji per phrase, and it remains future work to implement, in real-time, multiple tones into a single sentence similar to \cite{storytts}, but in a controllable manner. Moreover, in early explorations, it appeared that this effect was stronger for emojis that may focus on a particular word rather than on an entire phrase, e.g., a ``wink'' emoji. Future work should explore how emojis can be used as a form of markup for emphasis and tone around specific words in a phrase.

Interestingly, for both case studies, we had 3 participants who preferred the Baseline voice for the interactions, and stated that they prefer a flat, robotic sounding voice for a robot. This shows us that, despite the task, some individuals still want a classic, robotic voice for a robot embodiment. Moreover, all three of these individuals were male, and in general, the male participants seemed to provide lower scores to the expressive voices. Future work should be done to better understand gender preferences for expressive voices and tasks.

Moreover, as Miroka is still in early access, the participants have intimate knowledge of the robot. Therefore, there is a strong bias in testing new voices that, from the group discussions, likely resulted in reduced suitability scores.

Lastly, users expect clarity from the robot's speech. This can be aided by user adaptive and environmentally adaptive systems. Future work could be combined with a system such as \cite{renclarity} that updates the voice to use the user's language ability and environmental noise or utilizes known linguistic phenomena \cite{tuttosi24_interspeech} to increase clarity in words known to be difficult.

\section{CONCLUSIONS}
We have presented EmojiVoice, a toolkit for building expressive long-term varied-expressive speech systems using Matcha-TTS for robotics applications. Our toolbox guides social roboticists to create their own voices with as little as three minutes of data per voice, and through our case studies, we saw that the system runs in real-time. We further found that through the use of phrase-by-phrase emoji prompting, we were able to create a variable, expressive voice that improved impressions over a singularly expressive ``joyful'' voice in the longer storytelling use case. However, in a line-by-line assistant interaction, varied expressivity was not preferred. As such, researchers need to put care into considering all aspects of the task and interaction context when selecting and designing their robot voices.





\section*{ACKNOWLEDGMENT}

Mohammed Hafsati and Waldez Gomes for their help in running the Miroka experiments, Paul Maublanc for listening endlessly to our voices, Jean-Julien Aucouturier for his useful feedback, and Rajan Family for their support.


\bibliographystyle{IEEEtran}
\bibliography{IEEEexample}

\end{document}